# Intentional Deep Overfit Learning (IDOL): A Novel Deep Learning Strategy for Adaptive Radiation Therapy

**Running title:** Intentional Deep Overfit Learning

Jaehee Chun[3*], Justin C. Park[1*], Sven Olberg[1,2], You Zhang[1], Dan Nguyen[1], Jing Wang[1], Jin Sung Kim[3**], Steve Jiang[1]

[1]Medical Artificial Intelligence and Automation (MAIA) Laboratory, Department of Radiation Oncology, University of Texas Southwestern Medical Center, Dallas, TX 75390, USA

[2]Department of Biomedical Engineering, Washington University in St. Louis, St. Louis, MO 63110, USA

[3]Department of Radiation Oncology, Yonsei Cancer Center, Yonsei University College of Medicine, Seoul, South Korea

*Equal contribution co-first author

**Corresponding Author:

Jin Sung Kim, Ph.D.

Department of Radiation Oncology, Yonsei Cancer Center, Yonsei University College of Medicine, Seoul, South Korea

50, Yonsei-ro, Seodaemun-gu

Seoul, South Korea

T. +82-2-2228-8008, F.+82-2-2227-7823

Email: jinsung@yuhs.ac

**Conflicts of interest:** The authors have no conflicts to disclose.

**Acknowledgment:**

JS Kim acknowledges that this work was supported by the Korea Medical Device Development Fund grant funded by the Korean government (the Ministry of Science and ICT, the Ministry of Trade, Industry and Energy, the Ministry of Health & Welfare, Republic of Korea, the Ministry of Food and Drug Safety) (Project Number: 202012E01-03)

# ABSTRACT

**Purpose:** Applications of deep learning (DL) are essential to realizing an effective adaptive radiotherapy (ART) workflow. Despite the promise demonstrated by DL approaches in several critical ART tasks, there remain unsolved challenges to achieving satisfactory generalizability of a trained model in a clinical setting. Foremost among these is the difficulty of collecting a task-specific training dataset with high-quality consistent annotations for supervised learning applications. In this study, we propose a tailored DL framework for patient-specific performance that leverages the behavior of a model intentionally overfitted to a patient-specific training dataset augmented from the prior information available in an ART workflow – an approach we term Intentional Deep Overfit Learning (IDOL).

**Methods:** Implementing the IDOL framework in any task in radiotherapy consists of two training stages: 1) training a generalized model with a diverse training dataset of $N$ patients, just as in the conventional DL approach, and 2) intentionally overfitting this general model to a small training dataset-specific the patient of interest ($N + 1$) generated through perturbations and augmentations of the available task- and patient-specific prior information to establish a personalized IDOL model. The IDOL framework itself is task-agnostic and is thus widely applicable to many components of the ART workflow, three of which we use as a proof of concept here: the auto-contouring task on re-planning CTs for traditional ART, the MRI super-resolution (SR) task for MRI-guided ART, and the synthetic CT (sCT) reconstruction task for MRI-only ART.

**Results:** In the re-planning CT auto-contouring task, the accuracy measured by the Dice similarity coefficient improves from 0.847 with the general model to 0.935 by adopting the IDOL model. In the case of MRI SR, the mean absolute error (MAE) is improved by 40% using the IDOL framework over the conventional model. Finally, in the sCT reconstruction task, the MAE is reduced from 68 to 22 HU by utilizing the IDOL framework.

**Conclusions:** In this study, we propose a novel IDOL framework for ART and demonstrate its feasibility using three ART tasks. We expect the IDOL framework to be especially useful in creating personally tailored models in situations with limited availability of training data but existing prior information, which is usually true in the medical setting in general and is especially true in ART.

# 1. INTRODUCTION

Artificial intelligence (AI) systems, especially deep learning (DL)-based algorithms, have been explored for core components of the radiotherapy workflow including image enhancement, auto-segmentation, dose prediction, and auto-planning. State-of-the-art DL technologies have demonstrated superior performance compared to human-based or preceding computer-aided approaches through the self-taught discovery of informative representations and the use of hierarchical layers of learned abstraction. Given the current pace of advancement, DL and other AI-enabled technologies are poised to dramatically change the radiotherapy workflow.

Considering the time constraints faced in adaptive radiotherapy (ART), especially in an online ART workflow that takes place while the patient remains on-table in the treatment position, tasks in these settings are prime targets for DL-based automation. DL methods have been widely explored in the general radiotherapy setting,[1-3] but data availability is a limiting factor in a medical setting compared to applications in computer vision and other related fields. Under such low data regimes, achieving generalizability of the DL model will remain a fundamental challenge due to data availability, standardization, and the intra- and inter-observer variability inherent in some tasks.[4] Traditionally, investigators have framed the challenge as combatting overfitting – a scenario in which the model tends to "memorize" the training dataset but performs poorly when applied to unseen data not represented in the training sample. In an effort to prevent this, techniques such as dropout, batch normalization, data augmentation, and transfer learning have been widely utilized.[5-9] However, these techniques often fall short when the expressive power of the network is high, when the set of transformations (e.g. translation, rotation) utilized is limited, and when the entirety of the available prior knowledge is not leveraged.

To date, a unique characteristic among all medical applications of DL that is specific to ART has been entirely underutilized: there always exists prior knowledge for an ART patient. A given patient may have any combination of diagnostic scans that can be used in image-to-image translation tasks, for example, or planning images and contours that may be leveraged in the auto-contouring task. If this prior knowledge is incorporated into the training process of an ART-involved model, we propose that the performance of such a model should drastically improve when applied in the ART workflow for this patient precisely due to the behavior of a model that has been overfitted to patient-specific prior knowledge.

In this study, we propose a new paradigm shift away from combatting overfitting in radiotherapy settings. Conventional investigations dictate that a model applied in a clinical setting should generalize well to unseen patient data – this is uncontroversial and easily understood. Inspired by the patient-centric ethos of the era of precision medicine, we instead seek to leverage the behavior of a model that is intentionally overfitted using patient-specific prior information. Therefore, in what we term the Intentional Deep Overfit Learning (IDOL) approach, the challenge we are interested in is not to produce a model that generalizes well to all future patients but to optimize the performance of the model for this specific patient given the patient-specific prior information.

## 2. MATERIALS AND METHODS

### 2.1. Model

The learning framework used in this paper is a generalization of a semantic image translation/segmentation task consisting of a multi-task problem in which a set of $N$ functions must be approximated. Let $f = \{f_1, ..., f_N\}$ indicates the vector of functions to be learned. Let $X, Y$ be the input images and their counterparts in the true data space, where true data refers to existing data pairs for the desired task. Given the observation set $(\hat{X}_{train}, \hat{Y}_{train}) \subset (X, Y)$ as the training set, the goal is to find the inverse mapping function $f_\theta$ by optimizing a well-defined loss function $E$,

$$\hat{\theta} = \underset{\theta}{\mathrm{argmin}} \frac{1}{N} \sum_{(x,y) \in (\hat{X}_{train}, \hat{Y}_{train})}^{N} E(f_\theta(x), y). \tag{1}$$

The major risk in this framework is that when observations do not sufficiently cover the true data manifold, the resulting model generalizes poorly on unobserved data points. Here, given a separate validation set $(\hat{X}_{valid}, \hat{Y}_{valid}) \subset (X, Y)$, generalizability is quantified by the generalization error $E_{gen}$,[10] which may be formulated as

$$E_{gen} \cong \left| \frac{1}{N} \sum_{(x,y) \in (\hat{X}_{train}, \hat{Y}_{train})}^{N} E(f_\theta(x), y) - \frac{1}{M} \sum_{(x'',y'') \in (\hat{X}_{valid}, \hat{Y}_{valid})}^{M} E(f_\theta(x''), y'') \right|. \tag{2}$$

where the two terms of Eq (2) are the training error $E_{train}$ and validation error $E_{valid}$, respectively. When the training dataset is limited, $E_{train}$ converges to zero during the conventional DL training process while $E_{valid}$ saturates at a suboptimal level (Figure 1). In a robust DL model trained with data that sufficiently represents the characteristics of the unobserved validation data, $E_{gen}$ is ultimately small as $E_{train}$ and $E_{valid}$ converge to similar values.

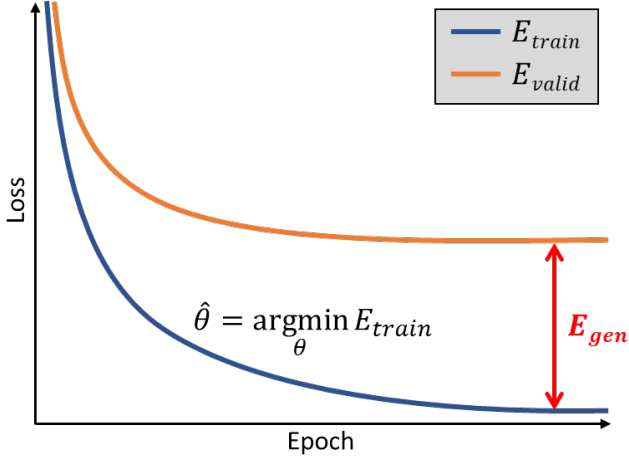

Figure 1. Characteristic training (blue) and validation (orange) curves illustrating loss values computed in each dataset during the conventional DL training process.

Traditional approaches utilize methods such as weight decay and dropout in efforts to minimize $E_{gen}$, however, it is fundamentally challenging to achieve this when the learned hypersurface does not effectively contain the latent space representation of unseen data as illustrated in Figure 2a.

In the proposed IDOL framework, we include a prior-knowledge mapping function $f' = \{f'_1, ..., f'_K\}$ from a set of $(\hat{X}_{prior}, \hat{Y}_{prior})$ that applies random deformation vector fields to a pair of task-specific prior data $(X_{prior}, Y_{prior}) \subset (X, Y)$ to produce augmented, patient-specific prior information. Following the conventional approach of constrained convex optimization, constraints can be enforced by penalizing the violation on the sample of training data together with another term that enforces the fitting of the supervised data to a prior. Mathematically, the proposed IDOL model can be formulated as

$$\hat{\theta} = \underset{\theta}{\text{argmin}}\{\lambda_l \frac{1}{N}\sum_{(x,y)\in(\hat{X},\hat{Y})}^{N} E(f_\theta(x), y) + \lambda_p \frac{1}{K}\sum_{(x',y')\in(\hat{X}_{prior},\hat{Y}_{prior})}^{K} \Theta(f_\theta(x'), y')\}. \quad (3)$$

Here $\Theta(f_\theta(x'), y')$ represents the penalty function for patient-specific prior information, $\lambda_l$ is the weight for the cost function optimization and $\lambda_p$ is the weight for the prior information term.

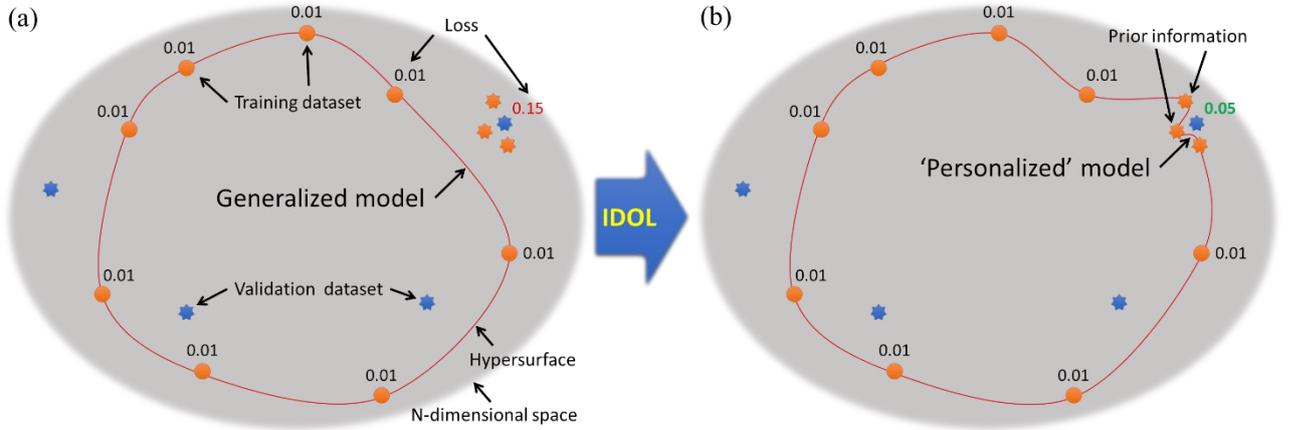

Figure 2. Conceptual representation of generalized DL (a) vs the proposed IDOL framework (b).

Figure 2 displays the conceptual representation of the generalized DL approach and our proposed IDOL framework. As Eq. (3) suggests, the training of IDOL comprises two steps. First, the generalized model is trained with a diverse set of $N$ patients with no prior knowledge of data to which the model will be applied, just as in the conventional DL approach (Figure 2a). In this case, validation samples may reside beyond the learned hypersurface in the latent space, giving rise to a higher computed loss in the validation set and the generalization error $E_{gen}$ observed in the conventional approach. Second, "intentional overfitting" is performed using augmentations of the available task- and patient-specific prior information to establish a personalized IDOL model characterized by a specifically molded learned hypersurface that yields improved, patient-specific performance for a given patient sample (Figure 2b). In the initial step, $E_{valid}$ becomes saturated at a certain point (Figure 1) due to the size of the training dataset and the associated poor generalization performance along with inherent limitations of the model architecture. In the second step in the IDOL framework, a sharp reduction in $E_{valid}$ is achieved owing to the introduction of a secondary training dataset consisting of patient- and task-specific augmentations of prior information (Figure 3). Here, we designate the improved generalization error as $E_{IDOL}$. The general workflow of the proposed IDOL framework is summarized in Figure 4.

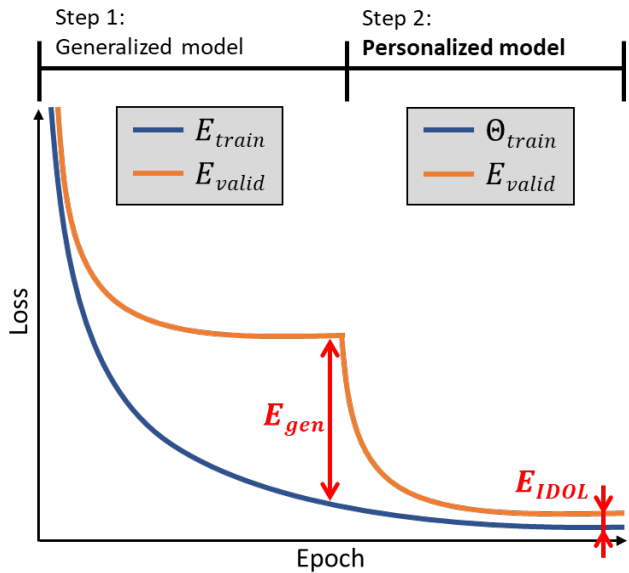

Figure 3. Characteristic training and validation curves of the IDOL approach demonstrating a significant improvement in patient-specific performance with the introduction of the second stage of training utilizing patient- and task-specific prior information.

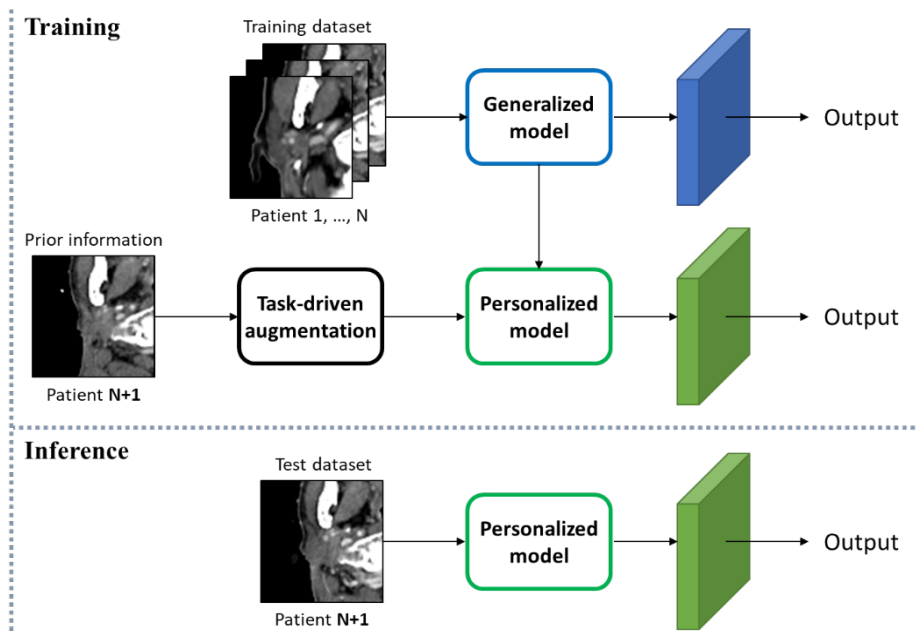

Figure 4. General workflow of the IDOL framework. Training consists of two steps. In the first step, a generalized model is produced utilizing a training dataset of the diverse patient dataset. Following this, task-specific augmentations of patient-specific prior information are utilized to further train the generalized model produced in step 1 and produce a personalized model for the patient of interest that may be applied at the time of inference.

## 2.2. Applications

The IDOL approach is a general framework that is widely applicable to various DL tasks. In this study, we focus on radiotherapy applications, especially those common to adaptive radiotherapy (ART) settings. We explore three such tasks here:

1) Auto-contouring/segmentation task on adaptive re-planning CT using initial planning CT / diagnostic CT as prior information.
2) MRI-Super Resolution based MRI Reconstruction for MRI-guided-Radiotherapy using simulation MRI / diagnostic MRI as prior information.
3) Synthetic-CT reconstruction from MRI using initial planning MRI-CT pairs as a prior information

### 2.2.1. Auto-contouring on re-planning CT

In adaptive settings, treatment re-planning is necessitated by changes in a patient's anatomy and tumor geometry from fraction to fraction. As one of the most time-intensive components of the ART workflow,[11] the re-contouring of organs at risk (OARs) and target structures on the re-planning CT (rpCT) is a prime target for automation through the use of DL auto-contouring approaches.[12] To assist with this process at the time of adaption, we utilize the IDOL framework with the planning CTs (pCTs) and contours drawn at the beginning of treatment serving as prior information.

In this experiment, 60 pCTs and manual contours (MCs) (patients P01-P60) were used as a training dataset for training a generalized auto-contouring model (step 1). In order to create a personalized model for each patient from the generalized model, pCTs and MCs of P61-P80 were used as the prior information in the repetition of step 2 for each patient. 20 rpCTs and MCs of P61-P80 that did not overlap with the training dataset were used as the final validation dataset.

### 2.2.2. Super-resolution on real-time MRI

Magnetic resonance image-guided radiotherapy (MR-IGRT) benefits from high-resolution (HR) acquisitions for tissue delineation and target tracking, but these HR images come at the cost of long scan times, reduced field of view, and reduced signal to noise ratio. Low-resolution (LR) acquisitions commonly acquired in the clinical setting to accommodate the reduced breath-hold capabilities of weaker patients can be enhanced through DL-based super-resolution (SR) approaches that reconstruct HR images from an LR input.[13] Personalized models for the MRI SR task can be deployed in real-time by utilizing LR and HR MRI pairs that are acquired for approximately 20 seconds before treatment as prior information.

100 pairs of LR and HR MRI for P81-111 were used to train a generalized SR model (step 1). Inhalation LR and HR MRI pairs from P112-115 were used to create a personalized model for each patient that was evaluated on a validation set of exhalation LR inputs and HR targets.

### 2.2.3. Synthetic CT generation from 3D MRI

An additional aspect of the adaptive MR-IGRT workflow is the challenge of multi-modality image registration in scenarios in which the anatomy observed in a simulation CT is incompatible with that observed in a setup MRI on the day of treatment. To overcome this challenge and avoid the acquisition of an rpCT, synthetic CT (sCT) images may be generated directly from MRI to meet the requirement of electron density information for dose calculations.[14] In the IDOL framework, simulation MR and CT images may be used as patient-specific prior information to drive sCT generation at subsequent treatment fractions.

The generalized sCT reconstruction model (step 1) was trained with data from a diverse population of P116-167 consisting of pairs of MRI and CT images (100 pairs per patient). Deformed simulation data (50 deformed sets per patient) of P168-170 were used to train a personalized model for each patient in step 2, which was validated in each case using setup MRI from the third treatment fraction for P168-170 (147 test images total).

## 3. RESULTS

### 3.1. Validation of the IDOL framework in the auto-contouring task

The auto-contouring results of the proposed IDOL framework are displayed in Figure 5a-f. It is evident that the auto-contour derived from the generalized model (Figure 5c) shows a large difference (Figure 5d) in terms of consistency compared to the target manual contour. In contrast, the IDOL model actively utilizes the prior information of a target patient's planning image and contours to generate contours with much higher consistency (Figure 5e-f).

Step 1 and Step 2 training was performed for 50 and 100 epochs, respectively, to ensure convergence of the loss function at each stage (Figure 6). The prediction accuracy, in terms of the Dice similarity coefficient (DSC), of the generalized model at the end of 50 epochs of training was $0.847 \pm 0.031$ on average for 20 patient cases. An additional 100 epochs of training in the IDOL framework saw an improvement in the average DSC score to $0.935 \pm 0.026$. Step 1 training took 20 minutes and Step 2 required an average of 20 seconds for each patient. The training and validation curves plotted throughout training (Figure 6) demonstrate the characteristic improvement in loss and accuracy with the introduction of Step 2 as illustrated in Figure 3.

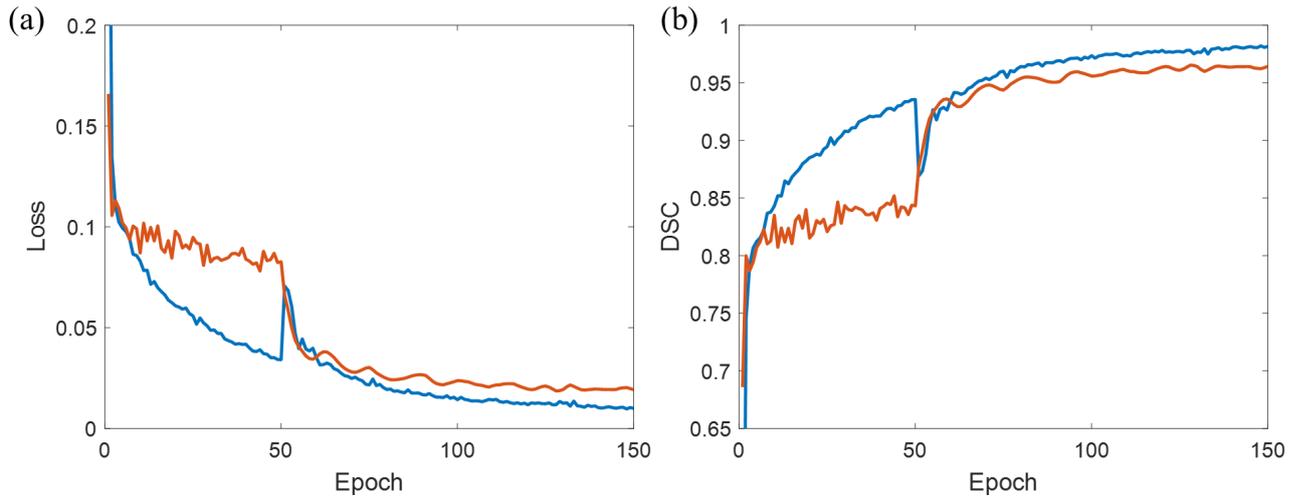

Figure 6. Training (blue) and validation (orange) curves in terms of loss value (a) and DSC (b) in the case of the auto-contouring task. Training of the generalized model was performed for 50 epochs, after which 100 epochs of patient-specific training were performed for each test patient.

### 3.2. Validation of the IDOL framework in the super-resolution task

MRI SR results are displayed in Figure 5g-l. Compared to the output of the generalized model (Figure 5i), the output of the patient-specific IDOL model (Figure 5k) is characterized by a clearer definition of the body contour and improved accuracy in the reconstruction of internal tissues.

As in the auto-contouring task, Step 1 and Step 2 training was performed for 50 (35 minutes) and 100 epochs (average 93 seconds per patient), respectively. The peak signal-to-noise ratio (PSNR) of SR images produced by the generalized model was averaged $30.69 \pm 2.15$ with a mean absolute error (MAE) of $4.18 \pm 0.86$, whereas the measures of the IDOL improved to $34.26 \pm 2.01$ and $2.50 \pm 0.54$, respectively.

### 3.3. Validation of the IDOL framework in the synthetic CT generation task

In the synthetic CT generation task explored here, outputs of the personalized model demonstrate improvements in boundary definition and tissue accuracy compared to outputs of the generalized model as illustrated in Figure 5m-r. Step 1 and Step 2 training was performed for 100 and 20 epochs, respectively. The MAE computed over 147 test images improved from $68 \pm 16$ HU with the general model to $22 \pm 5$ HU with the personalized model. Step 1 training took 21 hours and Step 2 training took an average of 3 hours for each patient.

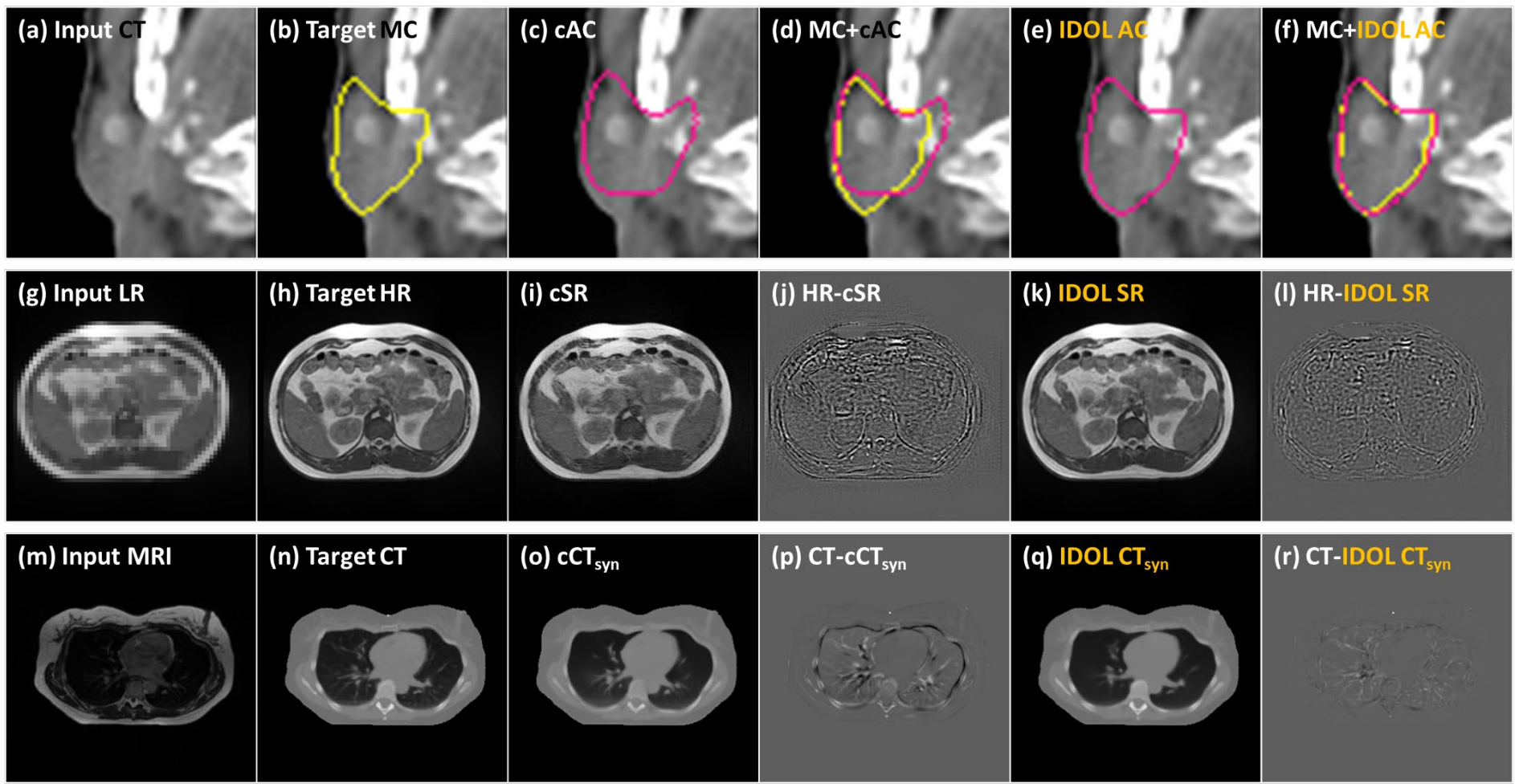

Figure 5. Results for the generalized model and patient-specific model trained in the IDOL framework in each of the three ART tasks explored here. For the auto contouring task on re-planning CTs (a-f), the target manual contour (MC) is shown along with the conventional auto-contour (cAC) produced by the general model and the IDOL AC. In the MRI super-resolution (SR) task (g-l), the input low-resolution (LR) and target high-resolution (HR) images are shown alongside the conventional SR (cSR) and IDOL SR outputs. For the synthetic CT reconstruction task (m-r), the input MRI and target CT are shown along with the outputs of the generalized and IDOL models.

## 4. DISCUSSION

The purpose of this study was to introduce a novel deep learning framework we term IDOL and demonstrate the successful application of the framework to common tasks in the ART workflow. In the auto-contouring task, the DSC accuracy improved by 0.088 from $0.847\pm0.031$ to $0.935\pm0.026$. In the MRI SR task, the MAE was reduced by over 40% from $4.18\pm0.86$ to $2.50\pm0.54$ with the adoption of the IDOL framework, which produces SR outputs with high fidelity to true HR images. A similar trend is observed in the synthetic CT generation task in the pixel-wise MAE is decreased by more than 65% from $68\pm16$ HU to $22\pm5$ HU. It is of note that the results here were achieved without any task-specific hyperparameter tuning. In this way, we anticipate that the IDOL framework is an extensible framework for patient-specific performance that is easily applied to any DL-based task in radiotherapy.

The main innovation of the IDOL approach can be described in two aspects. First, the adoption of the IDOL approach is a departure from the conventional wisdom in DL that overfitting is a primary challenge to overcome. Second, the intent of the IDOL framework is to achieve a high level of personalization as opposed to a high level of generalization, which could be described as the goal in the conventional case. To our knowledge, the IDOL approach is the first to utilize prior information successfully for a given patient to leverage the behavior of a model that has been intentionally overfitted and tailored for the patient- and task-specific performance. However, we must state that the investigations here are limited feasibility studies intended to act as a proof of concept. Therefore, our immediate next step is to evaluate the performance of the IDOL framework in tasks throughout the ART workflow for multiple sites, optimizing the approach and checking the stability of the framework in the clinical setting.

## 5. CONCLUSION

In this study, we have proposed a novel deep learning framework termed IDOL that utilizes intentional overfitting to develop patient-specific DL networks for common ART applications. The proposed IDOL concept is a global framework that is applicable to any task for which prior information exists, which is overwhelmingly the case in ART settings. Our IDOL approach embodies the ethos of the era of precision medicine in which personalized healthcare is the goal by foregoing interest in generalization to the next patient with optimal task performance for the present patient.

# REFERENCES

1. Cui S, Tseng HH, Pakela J, Ten Haken RK, El Naqa I. Introduction to machine and deep learning for medical physicists. *Med Phys*. Jun 2020;47(5):e127-e147. doi:10.1002/mp.14140
2. Luo Y, Chen S, Valdes G. Machine learning for radiation outcome modeling and prediction. *Med Phys*. Jun 2020;47(5):e178-e184. doi:10.1002/mp.13570
3. Seo H, Badiei Khuzani M, Vasudevan V, et al. Machine learning techniques for biomedical image segmentation: An overview of technical aspects and introduction to state-of-art applications. *Med Phys*. Jun 2020;47(5):e148-e167. doi:10.1002/mp.13649
4. Jarrett D, Stride E, Vallis K, Gooding MJ. Applications and limitations of machine learning in radiation oncology. *The British journal of radiology*. 2019;92(1100):20190001.
5. Srivastava N, Hinton G, Krizhevsky A, Sutskever I, Salakhutdinov R. Dropout: a simple way to prevent neural networks from overfitting. *The journal of machine learning research*. 2014;15(1):1929-1958.
6. Ioffe S, Szegedy C. Batch normalization: Accelerating deep network training by reducing internal covariate shift. PMLR; 2015:448-456.
7. Shorten C, Khoshgoftaar TM. A survey on image data augmentation for deep learning. *Journal of Big Data*. 2019;6(1):1-48.
8. Tan C, Sun F, Kong T, Zhang W, Yang C, Liu C. A survey on deep transfer learning. Springer; 2018:270-279.
9. Zhao Y, Rhee DJ, Cardenas C, Court LE, Yang J. Training deep-learning segmentation models from severely limited data. *Medical physics*. 2021;
10. Zhang C, Bengio S, Hardt M, Recht B, Vinyals O. Understanding deep learning requires rethinking generalization. *arXiv preprint arXiv:161103530*. 2016;
11. Henke L, Kashani R, Robinson C, et al. Phase I trial of stereotactic MR-guided online adaptive radiation therapy (SMART) for the treatment of oligometastatic or unresectable primary malignancies of the abdomen. *Radiotherapy and Oncology*. 2018;126(3):519-526.
12. Kim N, Chun J, Chang JS, Lee CG, Keum KC, Kim JS. Feasibility of Continual Deep Learning-Based Segmentation for Personalized Adaptive Radiation Therapy in Head and Neck Area. *Cancers*. 2021;13(4):702.
13. Chun J, Zhang H, Gach HM, et al. MRI super-resolution reconstruction for MRI-guided adaptive radiotherapy using cascaded deep learning: In the presence of limited training data and unknown translation model. *Medical physics*. 2019;46(9):4148-4164.
14. Olberg S, Zhang H, Kennedy WR, et al. Synthetic CT reconstruction using a deep spatial pyramid convolutional framework for MR-only breast radiotherapy. *Medical physics*. 2019;46(9):4135-4147.